\documentclass[sigconf]{acmart}
\AtBeginDocument{%
  }

\usepackage{multirow,multicol}
\usepackage{fancyhdr}
\renewcommand\footnotetextcopyrightpermission[1]{}
\setcopyright{acmlicensed}
\copyrightyear{2026}
\acmYear{2026}
\acmConference[Conference acronym 'MM]{}{}{}
\author{Hao-Yuan Ma}
\affiliation{%
  \institution{School of Computer Science and Technology, Soochow University}
  \city{Suzhou}
  \country{China}}

\author{Li Zhang}
\affiliation{%
  \institution{School of Computer Science and Technology, Soochow University}
  \city{Suzhou}
  \country{China}}

\author{Minjie Qiang}
\affiliation{%
  \institution{School of Computer Science and Technology, Soochow University}
  \city{Suzhou}
  \country{China}}

\author{Jie Gao}
\affiliation{%
  \institution{School of Computer Science and Technology, Soochow University}
  \city{Suzhou}
  \country{China}}

\renewcommand{\shortauthors}{Ma et al.}

\acmSubmissionID{668}

\begin{document}

\title{MambaCount: Efficient Text-guided Open-vocabulary Object Counting with Spatial Sparse State Space Duality Block}

\begin{abstract}
    Text-guided Open-vocabulary Object Counting (TOOC) aims to estimate the number of objects described by text prompts, which is particularly challenging in dense scenes with large scale variations. 
    Existing TOOC approaches predominately rely on Transformers, which quadratic complexity with respect to image resolution limits their scalability.
    Mamba offers a promising alternative due to its linear complexity.
    However, the previous Mamba methods have two main limitations. 
    On the one hand, the inherent causal formulation of Mamba constrains bidirectional spatial dependency modeling required by non-causal vision tasks. On the other hand, existing Mamba-based vision models often overlook the unconstrained high entropy in the spatial token, which can weaken local details and high-frequency cues.
    To address these limitations, we propose MambaCount, an efficient framework built on the Spatial Sparse State Space Duality block. Specifically, we analyze and reconstruct the decay dynamics of hidden states in Mamba to alleviate the dependency constraints introduced by causal modeling. Moreover, we introduce a Spatial Token Selection (STS) sub-block to reduce the unconstrained high entropy in spatial token responses within Mamba. In addition, we design Multi-Granularity Prototypes (MGP) to identify object-like regions at different semantic levels, improving cross-modal alignment and interpretability. Extensive experiments on FSC-147 demonstrate that MambaCount achieves state-of-the-art performance among methods without secondary querying, obtaining a test MAE of 12.23, while retaining linear complexity. The code is in the supplementary material.

\end{abstract}



\keywords{Object Counting, Vision-Language Model, State Space Duality, Prototype Learning}


\maketitle

\section{Introduction}
Unlike conventional class-specific counting \cite{FGENet,M2PLNet,vmambacc,CarPK} that is restricted to predefined categories such as crowds or vehicles.
Text-guided Open-Vocabulary Object Counting (TOOC) \cite{OVID-ma,CLIP-Count} aims to count objects of arbitrary categories specified by textual prompts, without visual exemplars. 
This flexibility makes TOOC particularly valuable in real-world applications, including inventory management in retail and warehousing \cite{warehouse}, ecological surveys to estimate wildlife populations \cite{wildlife}, and urban analytics for monitoring traffic and pedestrians \cite{urbanany}. Despite its practical significance, TOOC remains highly challenging due to the co-occurrence of large scale variations, dense and occluded distributions, and the open-ended nature of text-specified categories.

Existing TOOC methods can be broadly categorized into two paradigms. Firstly, exemplified by CLIP-Count \cite{CLIP-Count}, VLCounter \cite{VLCounter} and OVID \cite{OVID-ma}, employs CLIP-style \cite{CLIP} vision-language models which use standard Transformer and face the quadratic complexity $O(N^2)$ of the attention mechanism severely limits scalability. 
Secondly, represented by CountGD \cite{CountGD} and GroundingREC \cite{GroundingDINO}, builds upon GroundingDINO \cite{GroundingDINO} and adopts deformable attention \cite{defatten} to mitigate the cost of dense attention through sparse reference-point sampling. These methods typically impose a hard constraint on the maximum number of query points. When the object count exceeds this limit, they must resort to adaptive image cropping and secondary querying over local regions, resulting in substantial inference overhead \cite{CountGD}. Moreover, the computational advantage of deformable attention is not always guaranteed in practice. Under low-resolution inputs, where the number of visual tokens is relatively small, the overhead introduced by large query sets, multi-scale feature interaction, and deformable sampling can outweigh its sparsity benefits, sometimes even leading to higher computation than standard Transformers. In addition, these methods often rely on specialized operators to accelerate deformable attention, which further limits their practicality in real-world deployment. Overall, neither paradigm scales gracefully to the high-count, high-resolution scenarios common in real-world TOOC applications.


To overcome the above scalability bottlenecks, we turn to Mamba \cite{mamba,mamba2}, which offers linear $\mathcal{O}(N)$ complexity and has shown promising results in vision tasks. However, adapting existing Vision Mamba methods \cite{vmamba,vim} to object counting suffers from two key limitations: (1) their inherent causal nature necessitates 1D serialization, disrupting the non-causal 2D spatial structure of visual features; and (2) Mamba adaptations overlook the issue of unconstrained high entropy in spatial tokens, which can weaken local structures and suppress high-frequency cues that are essential for accurate object counting.
To address these deficiencies, we analyze and reconstruct the State Space Duality (SSD) framework in Mamba2 and propose the Spatial Sparse State Space Duality (S$^4$D) block for object counting. Specifically, we conduct an analysis of the decay dynamics of hidden states to alleviate the constraints of causal state transitions on non-causal spatial dependency modeling. In addition, we introduce a Spatial Token Selection (STS) sub-block to suppress unconstrained high entropy in spatial tokens, thereby preserving local details and discriminative high-frequency cues.


Furthermore, text-guided counting of arbitrary categories requires fine-grained cross-modal alignment between visual and textual representations. However, a single text embedding is often insufficient to align with visual patterns across different semantic granularities.
To alleviate this issue, we propose Multi-Granularity Prototypes (MGP), which derive fine-to-coarse prototypes from hierarchical feature maps through a top-$K$ selection strategy. By explicitly selecting the most text-relevant prototypes at each granularity, MGP enables more precise cross-modal alignment while suppressing background interference. In addition, the selected prototypes improve interpretability by revealing which image regions are regarded as object-like across different semantic granularities.

Our main contributions can be summarized as follows:
\begin{itemize}
    \item We propose the Spatial Sparse State Space Duality (S$^4$D) block. We analyze the State Space Duality (SSD) framework in Mamba2 and find that unconstrained high entropy in spatial tokens leads to weakening local details and high-frequency cues. We reconstruct the SSD framework to resolve the causal in vision tasks and introduce the Spatial Token Selection (STS) sub-block to mitigate the unconstrained high entropy in spatial token.
    \item We propose Multi-Granularity Prototypes (MGP), a hierarchical prototype learning module that extracts fine-to-coarse prototypes from multi-scale features. MGP provides fine-grained semantic alignment for accurate open-vocabulary counting and enhances model interpretability by explicitly revealing target-like regions at each granularity.
    \item We build MambaCount, a unified framework for Text-guided Open-Vocabulary Counting. Extensive experiments demonstrate that MambaCount achieves state-of-the-art (SOTA) performance among methods without secondary querying on FSC-147 with a Test MAE of 12.23, while maintaining linear computational complexity.
\end{itemize}
\section{Related Work}
\label{sec:related_work}

\subsection{Open-Vocabulary Object Counting}

The open-vocabulary object counting aims at estimating the number of target objects specified by flexible input prompts. Existing approaches can be broadly categorized according to the way the target object is specified.

\textbf{Exemplar-based Counting.} Early open-vocabulary counting methods, often referred to as few-shot counting, define the target object using visual exemplars such as bounding boxes or support patches. Representative works such as FamNet \cite{FamNet} and CFOCNet \cite{cfocnet} employ Siamese-style matching to compare the exemplar features with those of the query image. Although effective, these methods require manual annotation at inference time, which limits full automation.

\textbf{Text-guided Open-Vocabulary Counting.} Recent methods replace visual exemplars with textual prompts, enabling a more flexible and automated counting pipeline. Early approaches such as CLIP-Count \cite{CLIP-Count} and CounTX \cite{CounTX} leverage vision-language models to transfer semantic knowledge from text to density estimation. However, these methods typically inherit the dense attention computation of ViT-based backbones, resulting in quadratic complexity $\mathcal{O}(N^2)$ with respect to the number of visual tokens and limiting scalability on high-resolution images.
Another line of work adapts open-vocabulary detection models such as GroundingDINO \cite{GroundingDINO} for counting, including CountGD \cite{CountGD} and GroundingREC \cite{REC}. Although effective, these detector-based methods face notable scalability challenges. First, they are constrained by a fixed query budget (e.g., 900 object queries), which makes direct counting difficult in dense scenes. To handle high object counts, they often rely on image slicing and secondary querying over multiple sub-regions, substantially increasing inference cost. Second, their reliance on deformable attention introduces additional implementation overhead, since offset learning and sparse sampling can lead to irregular memory access and often require specialized operators for efficient execution. These limitations motivate the need for a more scalable and deployment-friendly solution for text-guided open-vocabulary counting.

\begin{figure}[t] 
    \centering
    \includegraphics[width=0.75\linewidth]{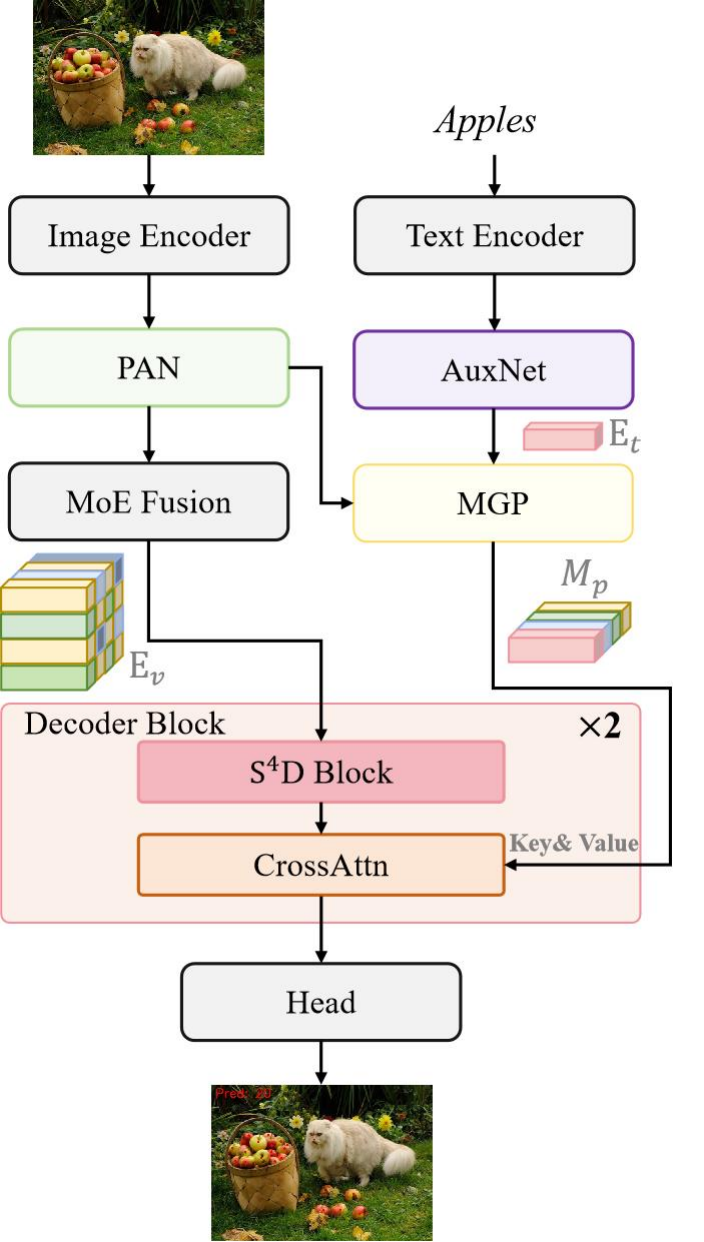}
    \caption{Overview of MambaCount. The framework consists of an image encoder and a text encoder. Visual features are enhanced by PAN and MoE Fusion to produce $E_v$, while textual features are refined by AuxNet to obtain $E_t$. MGP extracts fine-to-coarse prototypes $M_p$ from hierarchical features via top-$K$ selection. The decoder stacks S$^4$D block and cross-attention block, which use $M_p$ to guide cross-modal interaction. The prediction head outputs the counting result.} \vspace{-0.2cm}
    \label{fig:arch_overview}
\end{figure} \vspace{-0.2cm}

\subsection{Mamba in Vision}
Mamba \cite{mamba,mamba2} was originally developed for long-sequence modeling in natural language processing. Its main advantage lies in its linear computational complexity with respect to sequence length, offering a more efficient alternative to Transformers with quadratic complexity.

Motivated by its efficiency, recent studies have extended Mamba to the visual domain. Early works such as Vision Mamba (Vim) \cite{vim} and VMamba \cite{vmamba} introduced bidirectional scanning mechanisms to alleviate the directional bias of 1D state-space modeling, enabling efficient processing of high-resolution images. To further improve the ability to represent, Efficient Mamba \cite{efficientvmamba} incorporates convolutional attention and channel rotation to improve feature interaction. Similarly, DefMamba \cite{defmamba} adopts dynamic scanning strategies to reduce spatial information loss caused by fixed scanning paths.

Despite these advances, existing vision Mamba methods still face two fundamental limitations. First, many approaches rely on carefully designed scanning strategies to approximate non-causal spatial modeling with inherently causal scans. This not only deviates from the non-causal nature of vision tasks but may also depend on specialized implementations for efficient execution, which can limit deployment flexibility. Second, although recent work such as VSSD \cite{vssd} explores non-causal SSD formulations to relax the causality constraint, it does not explicitly address the unconstrained high entropy of spatial tokens. Such responses can weaken local details and high-frequency cues, which are particularly important for dense object counting.

In contrast, our method revisits the SSD formulation from the perspective of non-causal visual modeling and introduces a Spatial Sparse State Space Duality (S$^4$D) block for TOOC. By redesigning the hidden-state decay dynamics and incorporating Spatial Token Selection (STS), our approach improves non-causal spatial dependency modeling while better preserving local details and discriminative high-frequency information.

\section{Method}
\label{sec:method}

\subsection{Overview}

The overall architecture of MambaCount is illustrated in Fig.~\ref{fig:arch_overview}. Our framework follows a dual-encoder and cross-modal decoder paradigm, consisting of an image encoder, a text encoder, a Path Aggregation Network (PAN) \cite{PAN}, an Auxiliary Network (AuxNet), a Multi-Granularity Prototypes (MGP) module, a Mixture-of-Experts (MoE) fusion module, a decoder composed of stacked Spatial Sparse State Space Duality (S$^4$D) and cross-attention blocks, and a decoupled prediction head.

Given an input image and a textual query, the image encoder, extracts hierarchical visual features. These features are further enhanced by PAN to aggregate multi-scale spatial context, yielding visual representations $E_v^m,m\in\{1,2,3\}$. In parallel, the text encoder, instantiated with MobileCLIP \cite{mobileclip}, encodes the input text into semantic features, which are then refined by AuxNet to produce enriched textual embeddings $E_t$. Here, AuxNet is implemented as a lightweight Feed-Forward Network (FFN) with SwiGLU activation.

To improve cross-modal alignment across different semantic granularities, the MGP module extracts fine-to-coarse prototypes from hierarchical visual features through a top-$K$ selection strategy. The resulting prototype representations $M_p$ highlight object-like regions at different spatial scales and provide informative cues for text-guided matching, details are provided in sub-sec~\ref{sec:mgp}.

Next, the multi-scale visual features $E_v^m$ are adaptively integrated by the MoE fusion module to obtain fused visual representations. These fused features are then fed into the decoder, where each layer consists of an S$^4$D block followed by a cross-attention module. In the cross-attention stage, the prototype representations $M_p$ serve as keys and values to guide the cross-modal interaction with visual features.

Finally, the output of the decoder is fed into a decoupled prediction head to generate the counting result. Specifically, the prediction head contains two parallel convolutional branches, one for confidence prediction and the other for object localization.

\begin{figure}[h]
    \centering
    \includegraphics[width=0.85\linewidth]{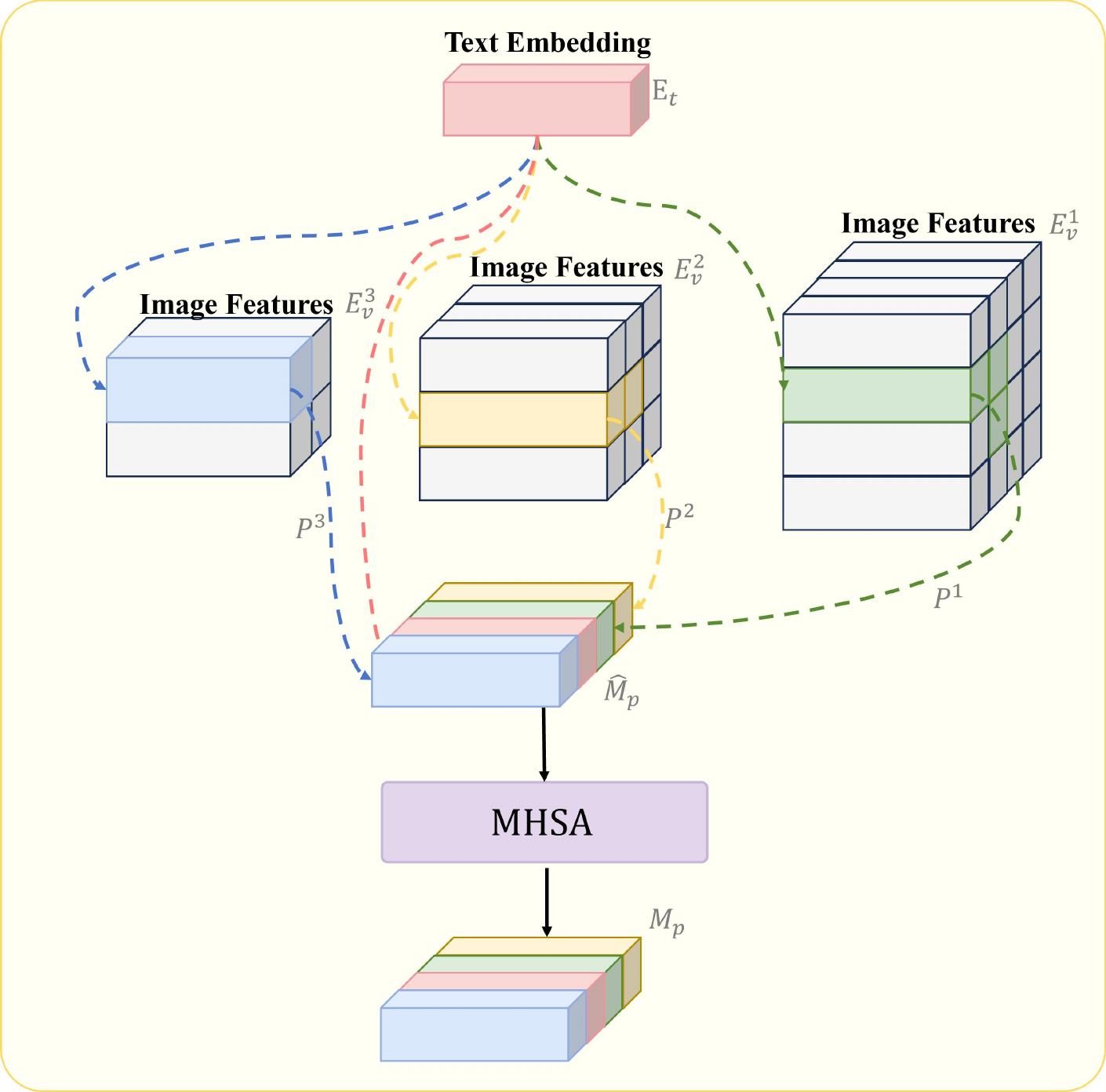}
    \caption{Illustration of the Multi-Granularity Prototypes (MGP). Given the text features $E_t$, we compute similarity with hierarchical image features and perform top-$K$ selection at each level to identify the most text-relevant regions. These selected tokens form multi-granularity prototypes representing object-like regions from fine to coarse semantic levels. The prototypes are further processed by Multi-Head Self Attention (MHSA) to capture their interactions and provide structured guidance for cross-modal reasoning.}
    \label{fig:mgp} \vspace{-0.2cm}
\end{figure}

\subsection{Multi-Granularity Prototypes} 
\label{sec:mgp}

TOOC requires precise semantic alignment between visual and textual representations. However, a single text embedding cannot simultaneously match visual semantics at different granularities. When biased toward high-level semantics, it tends to overlook fine-grained details, whereas focusing on low-level textures often introduces background noise. Moreover, computing text-image similarity over the entire feature map leads to global matching, which is particularly problematic in dense scenes where background regions may partially overlap with the target semantics, resulting in false positives.

To address these issues, we propose Multi-Granularity Prototypes (MGP) shown in Fig.~\ref{fig:mgp}, which explicitly extracts text-relevant visual prototypes from hierarchical feature maps. By selecting the most relevant regions at different semantic levels, MGP constructs fine-to-coarse representations that improve cross-modal alignment while suppressing background interference. Additionally, the selected prototypes enhance model interpretability by revealing which image regions are considered object-like at different spatial scales.

Given the visual encoder and PAN, we obtain multi-scale visual features:
\begin{equation}
E_v^m \in \mathbb{R}^{B\times C \times H_m \times W_m}, \quad m \in \{1,2,3\},
\end{equation}
where the spatial resolutions is defined as:
\begin{equation}
(H_1, W_1) = \left(\frac{H}{8}, \frac{W}{8}\right), \;
(H_2, W_2) = \left(\frac{H}{16}, \frac{W}{16}\right), \;
(H_3, W_3) = \left(\frac{H}{32}, \frac{W}{32}\right)
\end{equation}
where $H$ and $W$ denote the height and width of the input image resolution.
These hierarchical feature maps encode visual information at different semantic granularities. Meanwhile, the text encoder produces textual embeddings is given by:

\begin{equation}
E_t \in \mathbb{R}^{B\times K^p \times C},
\end{equation}

where $K^p$ denotes the number of textual tokens.

To identify text-relevant visual regions, we compute the similarity between text features and spatial visual features. First, the visual features are normalized can be calculated by:

\begin{equation}
\tilde{E}_v^m = \text{Norm}(E_v^m)
\end{equation}

The similarity between textual tokens and each spatial location of the visual feature map is computed through a channel-wise inner product. The relationship is expressed as:

\begin{equation}
\hat{S}^m_{b,k,h,w}
=
\left\langle 
\tilde{\mathbf{e}}^{\,m}_{v,b,h,w}, 
\mathbf{e}_{t,b,k}
\right\rangle
=
\sum_{c=1}^{C}
\tilde{E}^{m}_{v,b,c,h,w} \,
E_{t,b,k,c}
\end{equation}

where $b$ denotes the batch index, $k$ denotes the textual token index, and $(h,w)$ represents the spatial location.

To stabilize training, we introduce a learnable temperature parameter and bias, resulting in scaled similarity can be calculated by:

\begin{equation}
S^m_{b,k,h,w} =
\frac{\hat{S}^m_{b,k,h,w}}{\exp(-\alpha)} + \beta
\end{equation}

where $\alpha$ controls the temperature of the similarity distribution and $\beta$ is a learnable bias term.

Given the similarity map $S^m$, we select the top-$K$ spatial locations with the highest similarity scores. The calculation formula is given by:

\begin{equation}
\mathcal{I}^m = \text{TopK}(S^m, K).
\end{equation}

The corresponding visual tokens are extracted as prototypes can be calculated by:

\begin{equation}
P^m = \{E_v^m(i) \mid i \in \mathcal{I}^m\}.
\end{equation}

These prototypes represent the regions of objects similar to objects relevant to text at the $m$-th feature level.

Finally, prototypes from all feature levels are aggregated to form the multi-granularity prototype set:

\begin{equation}
\hat{M}_p = \text{Concat}(E_t,P^1, P^2, P^3).
\end{equation}

To capture interactions among these prototypes, we further apply a multi-head self-attention module:

\begin{equation}
M_p = \text{MHSA}(\hat{M}_p).
\end{equation}

The resulting prototypes highlight text-relevant regions across different spatial scales and serve as structured guidance for the subsequent cross-modal decoding stage.

\begin{figure}
    \centering
    \includegraphics[width=\linewidth]{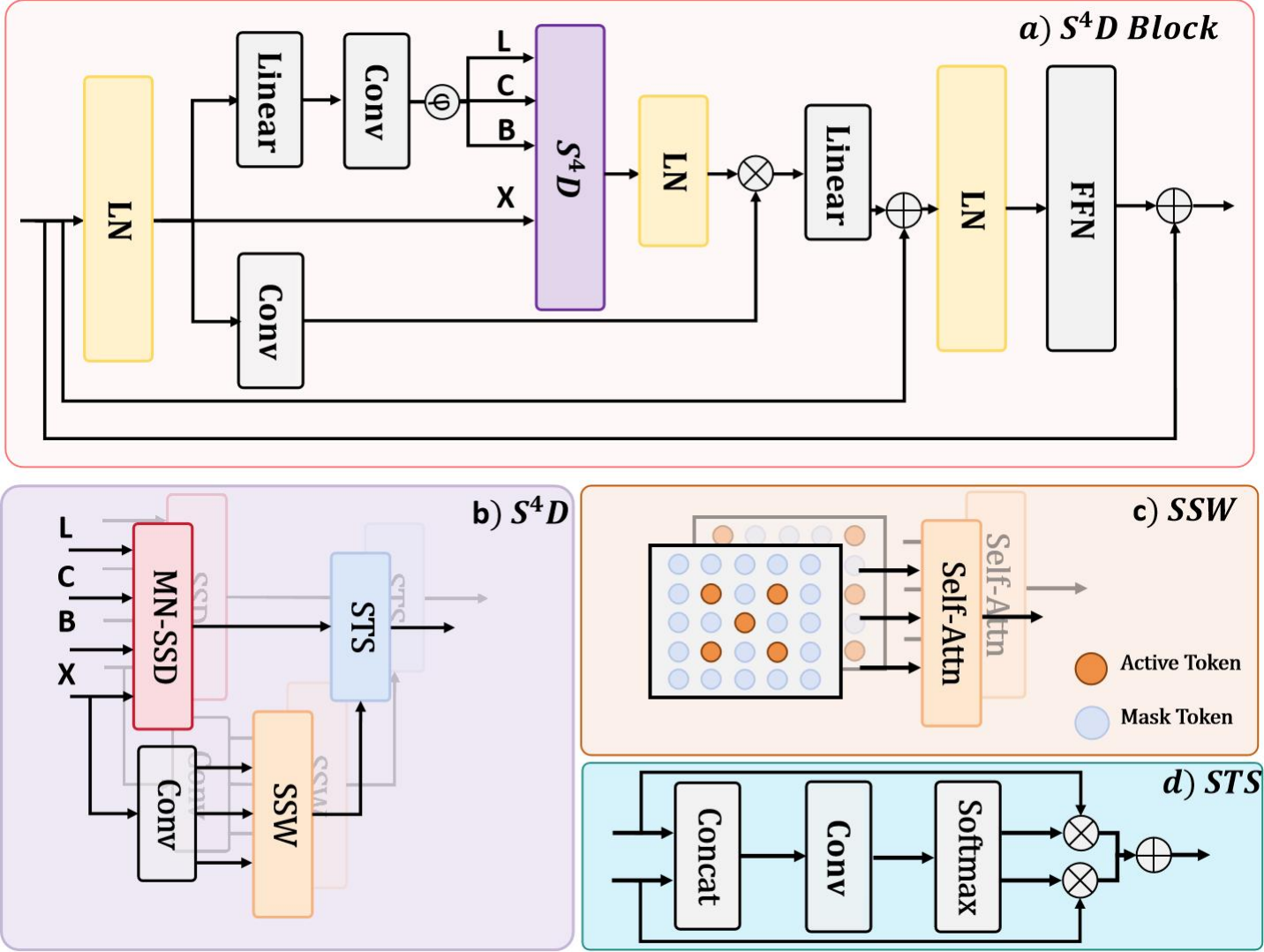}
    \caption{Overview of the proposed S$^4$D block.
(a) Overall structure of the S$^4$D block within the decoder. The input features are first normalized and projected to generate the parameters $(L, C, B)$ for the state space computation. The S$^4$D processes the sequence features and interacts with the convolutional branch through multiplicative gating, followed by a linear projection, an FFN, and residual connections.
(b) Internal design of the S$^4$D block, which combines the MN-SSD operator with two complementary components: the Spatial Sparse Window (SSW) sub-block and the Spatial Token Selection (STS) sub-block to model global dependencies and local structures.
(c) Structure of the SSW sub-block, where only a subset of spatial tokens are activated for self-attention while the remaining tokens are masked, enabling efficient sparse attention computation.
(d) Structure of the STS sub-block, which aggregates multi-branch features via concatenation and convolution operations, followed by softmax-based weighting to adaptively select informative spatial tokens.} \vspace{-0.2cm}
    \label{fig:s4d_decoder}
\end{figure} 
\subsection{Spatial Sparse State Space Duality Block} 
\label{sec:s4d}

Although Mamba-based architectures provide an efficient alternative to Transformers, directly applying existing vision Mamba blocks to dense object counting remains suboptimal. On the one hand, most existing methods inherit the causal dependency bias of state-space modeling and rely on scan-based serialization to process 2D feature maps. Such a design disrupts the native spatial structure of images and limits the modeling of bidirectional spatial interactions, which are crucial for counting objects. On the other hand, existing Mamba adaptations often produce overly diffuse, high-entropy spatial token responses, which weaken local details and high-frequency cues that are essential for distinguishing adjacent instances in counting task.

To address these limitations, we propose the Spatial Sparse State Space Duality (S$^4$D) block, as illustrated in Fig.~\ref{fig:s4d_decoder}(a). The S$^4$D block consists of a LayerNorm layer, an S$^4$D, a residual projection, and a feed-forward network. Among them, the S$^4$D, shown in Fig.~\ref{fig:s4d_decoder}(b), serves as the core component and is composed of three sub-blocks: a \emph{Multi-head Non-causal SSD (MN-SSD)} sub-block for global spatial modeling, a \emph{Spatial Sparse Window (SSW) attention} sub-block for local detail enhancement, and a \emph{Spatial Token Selection (STS)} sub-block for adaptively fusing the outputs of the two branches and alleviate the high-entropy spatial token responses.

Given an input feature map $X \in \mathbb{R}^{B\times H_2 \times W_2 \times C}$, we first apply layer normalization and feed the normalized features into two parallel branches. The global branch performs efficient long-range dependency modeling through the proposed MN-SSD, while the local branch captures fine-grained spatial structures via SSW attention. Their outputs are then adaptively integrated by the STS sub-block to form the output of the S$^4$D module. Finally, the fused features are projected, added to the input through a residual connection, and further refined by a feed-forward network, which produces the final output of the S$^4$D block.

\paragraph{\textbf{Multi-head Non-causal SSD (MN-SSD) sub-block}}
The first component of S$^4$D is a MN-SSD sub-block, which is designed to overcome the causal limitation of standard state-space modeling in vision. Unlike conventional Mamba-based blocks that depend on 1D scanning orders to approximate bidirectional modeling, our design performs non-causal global interaction directly on 2D visual features. 
In practice, the input features are projected to generate the parameters required by SSD, including the state transition and feature interaction terms for each head. 
\begin{equation}
B_h, C_h, X_h = W_B^h X, \; W_C^h X, \; X_h \in \mathbb{R}^{B\times N \times d_h}    
\end{equation}

where $N = H_2 \times W_2$ and $d_h = \frac{C}{H_a}$, and $H_a$ is the number of heads. 
Unlike Mamba2 which imposes a causal lower-triangular constraint on $\mathbf{L}$, we analyze the impact of causal mask and remove the causal mask, allowing each spatial token to attend to all other tokens bidirectionally. 
In the context of vision tasks, we reinterpret $\mathbf{L}$ as a spatially aware factor that captures positional relationships between tokens. In particular, the input $\mathbf{L}$ can be viewed as a form of two-dimensional positional encoding derived from the spatial layout of visual tokens. This spatial interpretation enables the model to incorporate structured positional information while preserving efficient state space computation.

\begin{align}
Y_{\mathrm{SSD}}^h &= \bigl( (\mathbf{B}_h+\mathbf{L})(\mathbf{C}_h+\mathbf{L})^\top \bigr)X_h \\
Y_{\mathrm{SSD}} &= Concat(Y_{\mathrm{SSD}}^h)
\end{align}

The general complexity of this branch is $\mathcal{O}(Nd_h^2)$ per head, linear in the number of spatial tokens.

These parameters are then used to model long-range spatial dependencies in a non-causal manner, allowing each spatial location to aggregate context from the entire feature map rather than from a single scan direction. In this way, the proposed Multi-head Non-causal SSD preserves the efficiency advantage of state-space modeling while being better aligned with the intrinsic non-causal nature of visual counting tasks.

\paragraph{\textbf{Spatial Sparse Window (SSW) attention sub-block}}



While the non-causal SSD branch is effective for global context modeling, dense object counting also requires fine-grained local structures to distinguish adjacent instances and preserve high-frequency details. To this end, we introduce the Spatial Sparse Window (SSW) attention sub-block, as shown in Fig.~\ref{fig:s4d_decoder}(c), to explicitly recover local details and high-frequency cues.

Inspired by dilated convolution, SSW constructs sparse window tokens by sampling local neighborhoods with different dilation rates, thereby enlarging the effective receptive field without introducing dense local attention. Specifically, we employ sparse window attention with multiple dilation rates $\mathcal{R}=\{1,3\}$, which provide effective receptive fields of $3\times3$ and $7\times7$, respectively. For each dilation rate $r \in \mathcal{R}$, the dilated neighborhood centered at position $(i,j)$ is defined as:

\begin{equation}
   \mathcal{N}_r(i, j) = \{(i + r \cdot \delta_h, \; j + r \cdot \delta_w) : \delta_h, \delta_w \in \{-1, 0, 1\}\} 
\end{equation}

where $\delta_h$ and $\delta_w$ denote the relative offsets in the vertical and horizontal directions, respectively, and $r$ controls the sampling interval of the sparse window.

Within each sparse window, self-attention is performed only on the sampled tokens. This operation is formulated as follows:

\begin{equation}
    \mathrm{SSW}_r(X)_{ij}
=
\sum_{(h,w)\in\mathcal{N}_r(i,j)}
\sigma\!\left(\frac{q_{ij}^{\top}k_{hw}}{\sqrt{d_h}}\right)v_{hw}    
\end{equation}

where $\sigma$ denotes the Softmax.

The outputs from different dilation rates are then aggregated as:
\begin{equation}
Y_{\mathrm{SSW}}
=
\frac{1}{|\mathcal{R}|}
\sum_{r\in\mathcal{R}}
\mathrm{SSW}_r(X).
\end{equation}

Compared with the local convolution adopted in standard Mamba-style blocks, SSW performs sparse 2D local modeling with multi-scale receptive fields, making it more suitable for capturing fine-grained textures, local structures, and geometric patterns in dense counting scenes.






\paragraph{\textbf{Spatial Token Selection}}
To dynamically allocate SSD or SSW modeling per spatial location, we introduce a Spatial Token Selection (STS) sub-block, as illustrated in Fig.~\ref{fig:s4d_decoder}(d). Different with the temporal selection mechanism $\Delta = \text{softplus}(\text{Linear}(x))$ in Mamba, but operates in the spatial domain.

The gate predicts a mask per-location $M \in [0, 1]^{H \times W}$ from the input features is defined as:

\begin{equation}
    M=\sigma(\frac{\Phi([Y_{SSD},Y_{SSW}])}{\tau})
\end{equation}

where $\Phi$ is a lightweight depthwise-separable convolution, and $\tau = |\theta_\tau| + 0.1$ is a learnable temperature parameter controlling the sharpness of the gate. The final S$^4$D output is:

$$Y_{S^4D} = M \circ Y_{SSD} + (1 - M) \circ Y_{SSW}$$

Where $\circ$ is the hadamard product. When $M \to 1$, the position relies on global context through MN-SSD, which is beneficial for sparse regions requiring long-range dependencies. When $M \to 0$, the position uses sparse window attention, which is more effective for dense regions where local features are sufficient for discrimination. Based on STS sub-block, the $S^4D$ block can constrain the high entropy in spatial token.

\begin{table*}[t]
\centering
\caption{Comparison results with existing methods on FSC147 dataset.}
\label{tab:main_results}
\resizebox{0.85\linewidth}{!}{
\begin{tabular}{lccc|cc|cc}
\hline
\multirow{2}{*}{Method} & \multirow{2}{*}{Type} & \multirow{2}{*}{Backbone} & \multirow{2}{*}{Decoder} 
& \multicolumn{2}{c|}{Val set} & \multicolumn{2}{c}{Test set} \\
\cline{5-8}
 &  &  &  & MAE $\downarrow$ & RMSE $\downarrow$ & MAE $\downarrow$ & RMSE $\downarrow$ \\
\hline
ZSC \cite{zsc} & Density & VAE & Transformer & 26.93 & 88.63 & 22.09 & 115.17 \\
CLIP-Count \cite{CLIP-Count} & Density & CLIP & Transformer & 18.79 & \textbf{61.18}& 17.78 & 106.62 \\
VLCount \cite{VLCounter} & Density & CLIP & Transformer & 18.06 & 65.13 & 17.05 & \textbf{106.16} \\
CountTX \cite{CounTX} & Density & OpenCLIP & Transformer & \textbf{17.33} & 61.58 & \textbf{16.28} & 106.41 \\ \hline
OVID \cite{OVID-ma} & Point & CLIP & Transformer & 17.87 & 64.18 & 15.61 & 98.73 \\ \hline
COUNTGD \cite{CountGD} & Detection & GDino & DefTransformer & 12.14 & 47.51 & 12.98 & \textbf{98.35} \\
GroundingREC \cite{REC} & Detection & GDino & DefTransformer & 10.06 & 58.62 & 10.12 & 107.19 \\
TrueCount$_{txt}$ \cite{truecount} & Detection & GDino & DefTransformer & \textbf{9.54} & \textbf{48.01} & 10.11 & 106.11 \\
DCount \cite{DCount} & Detection & GDino & DefTransformer & 9.63 & 53.60 & \textbf{9.83} & 105.15 \\
\hline
Ours & Point & YOLOE & Mamba & \textbf{13.22} & \textbf{53.62} & \textbf{12.23} & \textbf{105.69} \\
\hline
\end{tabular}
}
\end{table*}

\subsection{Training Objectives}

Inspired by P2PNet \cite{P2PNet}, the overall training objective combines classification and localization losses based on a strictly one-to-one label assignment. We first utilize the Hungarian algorithm \cite{kuhn1955hungarian} to match the predicted points with the ground-truth points. 

For category classification, we apply the cross-entropy loss on the confidence predictions from the decoupled head across all points:
\begin{equation}
    \mathcal{L}_{cls} = \text{CrossEntropy}(\hat{c}, c^*)
\end{equation}
where $\hat{c}$ denotes the predicted confidence scores and $c^*$ denotes the assigned ground-truth labels. 

For object localization, we adopt the smooth regression loss in OVID \cite{OVID-ma} to supervise coordinate regression. This loss is applied exclusively to the positively matched foreground points:
\begin{equation}
    \mathcal{L}_{loc} = \mathcal{L}_{
Smooth_{ln}}(\hat{p}, p^*)
\end{equation}
where $\hat{p}$ and $p^*$ are the predicted positions of the ground-truth points of the pairs matched, respectively. 

The total loss is computed as:
\begin{equation}
    \mathcal{L} = \mathcal{L}_{cls} + \lambda \mathcal{L}_{loc}
\end{equation}
where $\lambda$ is a hyperparameter designed to balance the two loss terms.

\section{Experiment}
\subsection{Experiment Settings}
\paragraph{\textbf{Dataset}}
We evaluate our MambaCount on three distinct benchmarks, each selected to validate specific capabilities of our framework. FSC-147 \cite{FamNet}, comprising 147 categories, serves as our primary benchmark to evaluate the core text-guided open-vocabulary counting ability of the model. To assess the zero-shot generalization and robustness of the model in handling extreme density, we tested it on CARPK \cite{CarPK}, which features densely packed cars captured from drone views. Furthermore, we evaluated on REC-8K \cite{REC} to demonstrate our model's superiority in addressing the challenges of referring expression.




\paragraph{\textbf{Implementation details}}
We adopt YOLOv11s \cite{yolov11} from the YOLOE \cite{yoloe} framework as the visual backbone to extract image features, and we employ MobileCLIP-BLT \cite{mobileclip} as the text encoder to obtain textual representations and freeze the text encoder. The model is trained on the FSC147 dataset for 200 epochs with a batch size of 8 and a random seed of 42. The $\lambda$ in loss is designed to 0.05. The learning rate is set to 1e-4 for the fusion module and 1e-5 for the remaining modules. We use the Adam \cite{adam} optimizer for training. All experiments are conducted on a server equipped with an NVIDIA RTX V100 GPU (32GB).





\subsection{Comparison results with Other Methods on three TOOC Datasets}

\paragraph{\textbf{Comparison results with existing methods on the FSC147 dataset}}
Table~\ref{tab:main_results} compares MambaCount with existing text-guided counting methods on FSC-147. Compared with density-based methods, MambaCount reduces the Test MAE from 16.28 (CountTX) to 12.23, showing the advantage of explicit point prediction over density regression. It also outperforms the point-based method OVID (15.61) by a clear margin, demonstrating the effectiveness of our Mamba-based vision-language fusion.
Compared with detection-based methods, MambaCount remains competitive. It improves over CountGD (12.98), while being slightly inferior to stronger methods such as GroundingREC, TrueCount$_{txt}$, and DCount. Notably, some of these methods rely on stronger detection pipelines or additional auxiliary priors, such as SAM \cite{segmentanyhing} and Depth Anything V2 \cite{depthv2} in TrueCount$_{txt}$. In contrast, MambaCount adopts a simpler point prediction framework without extra auxiliary modules, yet still achieves competitive performance.

\begin{table}[t]
\centering
\caption{Comparison results on the REC-8K dataset.}
\label{tab:comparison}
\resizebox{1\linewidth}{!}{
\begin{tabular}{l|cc|cc}
\hline
\multirow{2}{*}{Method} & \multicolumn{2}{c|}{Val Set} & \multicolumn{2}{c}{Test Set} \\
\cline{2-5}
 & MAE $\downarrow$ & RMSE $\downarrow$ & MAE $\downarrow$ & RMSE $\downarrow$ \\
\hline
Mean & 14.28 & 27.75 & 13.75 & 25.91 \\
ZSC$_{ResNet50}$ \cite{zsc} & 14.84 & 31.30 & 14.93 & 29.72 \\
ZSC$_{SwinT}$ \cite{zsc} & 12.96 & 26.74 & 13.00 & 29.07 \\
CountTX \cite{CounTX} & 11.88 & 27.04 & 11.84 & 25.62 \\
GroundingDINO$_{SwinT}$ \cite{GroundingDINO} & 9.03 & 21.98 & 8.88 & 21.95 \\
GroundingREC$_{SwinT}$ \cite{REC} & 6.80 & 18.13 & 6.50 & 19.79 \\
CAD-GD$_{SwinT}$ \cite{CAD-GD} & 5.43 & \underline{15.01} & \underline{5.29} & \underline{17.08} \\
CAD-GD$_{SwinB}$ \cite{CAD-GD} & \textbf{4.23} & \textbf{13.14} & \textbf{4.34} & \textbf{12.93} \\
DCount$_{SwinT}$ \cite{DCount} & \underline{5.24} & 15.47 & 5.45 & 18.61 \\
\hline
Ours & 5.28 & 16.26 & 5.42 & 17.18 \\
\hline
\end{tabular}
}
\end{table}
\paragraph{\textbf{Comparison results on REC-8K dataset}}
Table~\ref{tab:comparison} reports the comparison on the REC-8K dataset. Although MambaCount is not specifically designed for referring expression counting, it achieves strong performance with 5.42 MAE and 17.18 RMSE on the test set. Compared with early zero-shot and density-based methods such as ZSC and CountTX, our method reduces the counting error by a large margin. It also outperforms GroundingDINO and GroundingREC, and achieves performance comparable to recent strong detection-based methods such as CAD-GD$_{SwinT}$ and DCount$_{SwinT}$. These results demonstrate the strong generalization ability and competitiveness of MambaCount without any task-specific modifications.

\begin{table}[t]
\centering
\caption{Comparison results on the CARPK dataset}
\label{tab:carpk_results}
\begin{tabular}{lccc}
\hline
Dataset & Method & MAE $\downarrow$ & RMSE $\downarrow$ \\
\hline
\multirow{5}{*}{CARPK}
 & CLIP-count \cite{CLIP-Count} & 11.96 & 16.61 \\
 & CountTX \cite{CounTX} & 6.13 & 10.87 \\
 & VLCounter \cite{VLCounter} & 6.46 & 8.68 \\
 & CountGD \cite{CountGD} & \textbf{3.83} & \textbf{5.41} \\ \hline
 & Ours & 4.31 & 7.42 \\
\hline
\end{tabular}
\end{table}

\paragraph{\textbf{Results compared with existing SOTA methods on the CARPK dataset}}
Table~\ref{tab:carpk_results} presents the comparison with existing methods on the CARPK dataset. Our method achieves 4.31 MAE and 7.42 RMSE, outperforming most vision-language counting approaches and approaching the performance of the best detection-based model.
Compared with density regression methods such as CLIP-count, CountTX, and VLCounter, our method significantly reduces the counting error. In particular, the Test MAE decreases from 6.13 (CountTX) and 6.46 (VLCounter) to 4.31, demonstrating that explicit object-level modeling provides more accurate counting than density estimation in structured scenes such as parking lots.
CountGD achieves the best performance (3.83 MAE) due to the strong detection prior from large-scale grounding models. Nevertheless, our method achieves competitive performance without relying on heavy detection pipelines, indicating that the proposed framework can effectively capture object-level representations while maintaining a simpler architecture.

\begin{table}[t]
\centering
\caption{Ablation study on each component of MambaCount.}
\label{tab:ablation_module}
\resizebox{1\linewidth}{!}{
\begin{tabular}{l|cc|cc}
\hline
\multirow{2}{*}{} & \multicolumn{2}{c|}{Val set} & \multicolumn{2}{c}{Test set} \\
\cline{2-5}
 & MAE $\downarrow$ & RMSE $\downarrow$ & MAE $\downarrow$ & RMSE $\downarrow$ \\
\hline
Baseline & 15.81 & 64.82 & 16.07 & 114.14 \\
Baseline + S$^4$D Block & 14.66 & 56.41 & 16.09 & 110.84 \\
Baseline + MGP & 13.51 & 55.46 & 16.10 & 115.85 \\
All & \textbf{13.22} & \textbf{53.62} & \textbf{12.23} & \textbf{105.69} \\
\hline
\end{tabular}
}
\end{table}

\subsection{Ablation Experiment}

\begin{figure*}
    \centering
    \includegraphics[width=0.8\linewidth]{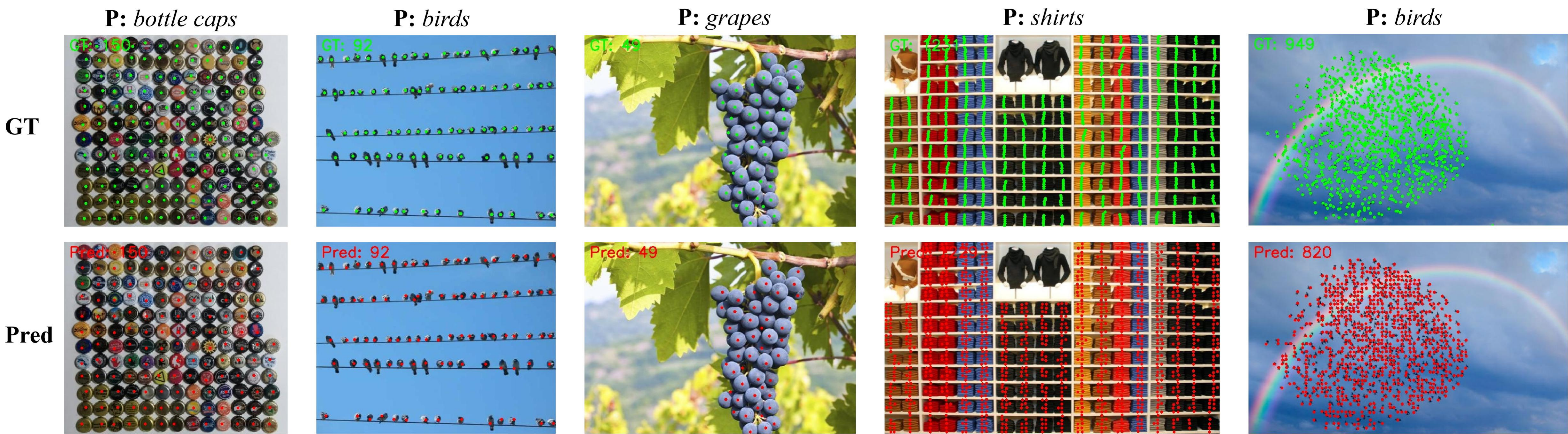}
    \caption{Visualization of the MambaCount on FSC147 dataset.}
    \label{fig:res_vis}
\end{figure*}
\paragraph{\textbf{Component analysis}}
Table~\ref{tab:ablation_module} shows the contribution of each component in MambaCount. Both S$^4$D Block and MGP Block improve the validation performance compared with the baseline. While each individual module brings limited improvement on the test set, combining them yields the best overall results, reducing the Test MAE from 16.07 to 12.23. This demonstrates that the two modules are complementary and both are important for the final performance.

\begin{table}[t]
\centering
\caption{Ablation study of MGP modules and comparison with Soft Exemplar (SE).}
\label{tab:ablation_mpg}
\resizebox{1\linewidth}{!}{
\begin{tabular}{l|cc|cc}
\hline
\multirow{2}{*}{Method} & \multicolumn{2}{c|}{Val set} & \multicolumn{2}{c}{Test set} \\
\cline{2-5}
 & MAE $\downarrow$ & RMSE $\downarrow$ & MAE $\downarrow$ & RMSE $\downarrow$ \\
\hline
Baseline & 14.66 & 56.41 & 16.09 & 110.84 \\
Baseline + SE & 13.51 & 55.46 & 16.10 & 115.85 \\
Baseline + MGP & \textbf{13.22} & \textbf{53.62} & \textbf{12.23} & \textbf{105.69} \\
\hline
CountTX & 17.33 & 61.58 & 16.28 & 106.41 \\
CountTX + SE & 17.02 & 62.98 & 16.29 & 106.02 \\
CountTX + MGP & \textbf{16.13} & \textbf{58.62} & \textbf{15.31} & \textbf{104.95} \\
\hline
\end{tabular}
}
\end{table}

\paragraph{\textbf{Impact of the MGP}}
Table~\ref{tab:ablation_mpg} evaluates the contributions of the Soft Exemplar (SE) \cite{countse} and MGP modules. Introducing SE provides a slight improvement on the validation set (MAE 14.66 $\to$ 13.51), but shows limited gains on the test set. In contrast, incorporating MGP significantly improves performance, reducing the Test MAE from 16.09 to 12.23 on the baseline and from 16.28 to 15.31 when applied to CountTX. These results demonstrate that the proposed MGP effectively enhances vision-language alignment and improves counting accuracy.

\begin{table}[t]
\centering
\caption{Ablation study of the $S^4D$ Block.}
\label{tab:attention_ablation}
\resizebox{0.9\linewidth}{!}{
\begin{tabular}{l|cc|cc}
\hline
\multirow{2}{*}{Module} & \multicolumn{2}{c|}{Val set} & \multicolumn{2}{c}{Test set} \\
\cline{2-5}
 & MAE $\downarrow$ & RMSE $\downarrow$ & MAE $\downarrow$ & RMSE $\downarrow$ \\
\hline
SSD & 14.27 & 60.68 & 14.82 & 112.48 \\
MN-SSD & 13.42 & 61.84 & 14.61 & 112.04 \\
SSW & 13.79 & 58.08 & 15.96 & 127.65 \\ \hline
$S^4D$ Block & \textbf{13.22} & \textbf{53.62} & \textbf{12.23} & \textbf{105.69} \\
\hline
\end{tabular}
}
\end{table}

\paragraph{\textbf{Impact of the $S^4D$ Block}}
Table~\ref{tab:attention_ablation} compares different spatial modeling mechanisms. The proposed $S^4D$ consistently outperforms SSD \cite{mamba2}, MN-SSD and SSW, achieving the best MAE and RMSE on both validation and test sets. This confirms that the proposed spatial modeling strategy provides more effective feature aggregation for counting tasks.

\begin{table}[htbp]
\centering
\caption{Comparison of different methods on $384 \times 384$ images with batch size = 1.}
\label{tab:efficiency_384}
\begin{tabular}{lccc}
\hline
Method & Param $\downarrow$ & FLOPs $\downarrow$ & FPS $\uparrow$ \\
\hline
CountTX                 & 161.08M & 43.88G & 3.2  \\
CountGD                 & 213.00M & 98.96G & 1.3  \\
CountGD (AC) & 213.00M & -       & 0.3  \\
MambaCount              & 41.77M  & 27.50G & 24.3 \\
\hline
\end{tabular}
\end{table}


\subsection{Efficiency Analysis}
Table~\ref{tab:efficiency_384} shows the efficiency comparison on $384 \times 384$ images with a batch size of 1. 
MambaCount achieves the best efficiency, with the fewest parameters (41.77M), lowest FLOPs (27.50G), and fastest inference speed (24.3 FPS). 
In contrast, CountTX and CountGD require higher computational cost and run much slower. 
After introducing Adapative Crop (AC), the inference speed of CountGD further drops to 0.3 FPS, indicating additional computational overhead.

\subsection{Visualization}
To further understand the effectiveness of the proposed design, we provide visual analysis from four perspectives. 
The Fig~\ref{fig:res_vis} presents the prediction results, where the proposed method produces object responses that are more accurate and spatially aligned with the targets, especially in dense, cluttered, and scale-varying scenes. 
The predicted results are more compact and better distributed around true object instances, indicating stronger counting reliability and localization consistency.

\begin{figure}
    \centering
    \includegraphics[width=0.85\linewidth]{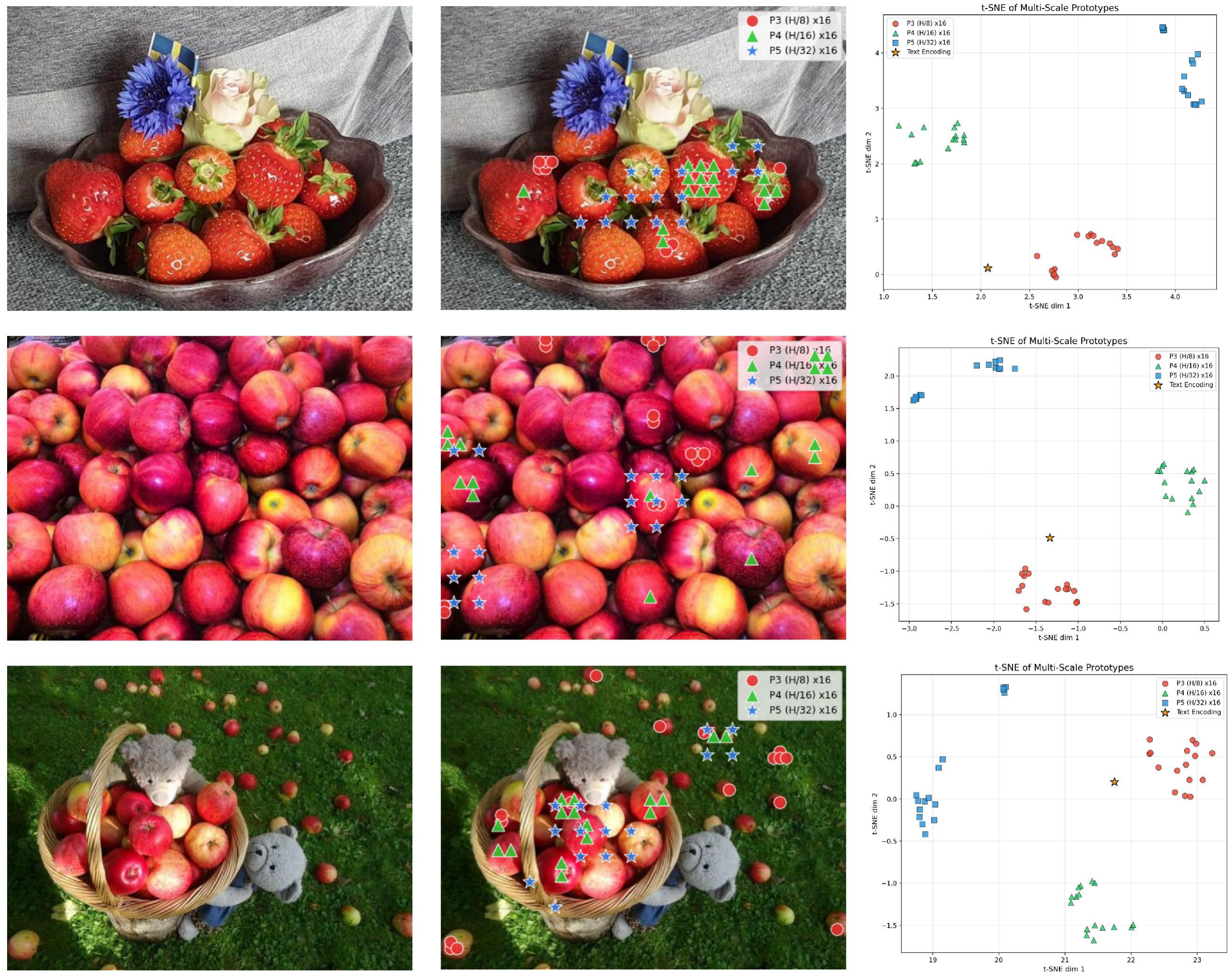}
    \caption{Visualization of the MGP. Prototypes are extracted from three different scales to capture complementary multi-scale representations. The t-SNE plots show clear separability among prototypes from different scales.}
    \label{fig:vis_mpg}
\end{figure}
The Fig~\ref{fig:vis_mpg} shows the multi-granularity prototypes (MGP). 
As shown in the prototype assignment and t-SNE plots, prototypes from different scales exhibit clear semantic separation while maintaining complementary relationships. 
Fine-scale prototypes mainly respond to small and dense instances, whereas coarser-scale prototypes capture larger structures and broader contextual patterns. 
This confirms that MGP provides discriminative multi-scale representations and improves adaptability to large variations in object size and arrangement.

\begin{figure}
    \centering
    \includegraphics[width=0.9\linewidth]{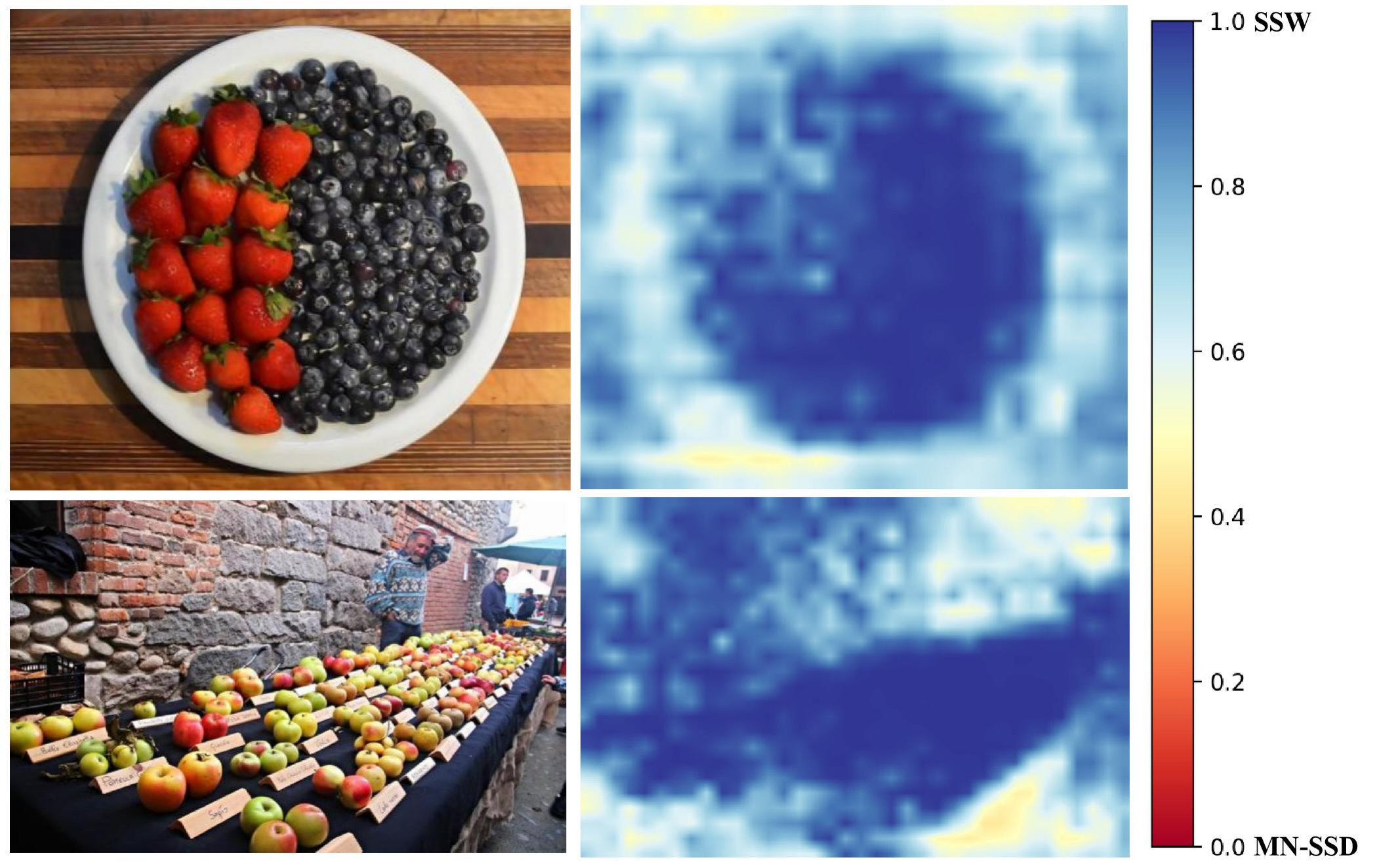}
    \caption{Visualization of the STS sub-block. STS combines SSW and MN-SSD to adaptively focus on informative regions. The resulting heatmaps demonstrate its ability to handle different object layouts.}
    \label{fig:vis_sts}
\end{figure}
The Fig~\ref{fig:vis_sts} shows the visual gating behavior of STS over SSW and MN-SSD. 
Specifically, STS assigns stronger local gating responses to SSW in regions with concentrated object details, which helps preserve fine-grained local structure. 
Meanwhile, MN-SSD receives stronger global responses over spatially broader regions, enabling more effective modeling of long-range dependencies and holistic scene layouts. 
Such complementary gating behavior demonstrates that STS can adaptively balance local detail aggregation and global context modeling and alleviates the high entropy in spatial token according to the image content.

\section{Conclusion}
In this paper, we propose MambaCount, an efficient framework for text-guided open-vocabulary object counting (TOOC) built upon the Spatial Sparse State Space Duality (S$^4$D) block. Our approach effectively addresses the limitations of applying Mamba to vision tasks.
Furthermore, the proposed Multi-Granularity Prototypes (MGP) enhance semantic alignment and improve the interpretability of the counting process. Extensive experiments on FSC-147 demonstrate that MambaCount achieves state-of-the-art performance among methods without secondary querying while maintaining linear computational complexity, highlighting its efficiency and scalability in TOOC.


\bibliographystyle{ACM-Reference-Format}
\bibliography{acmart}

\appendix
\begin{figure*}
    \centering
    \includegraphics[width=0.85\linewidth]{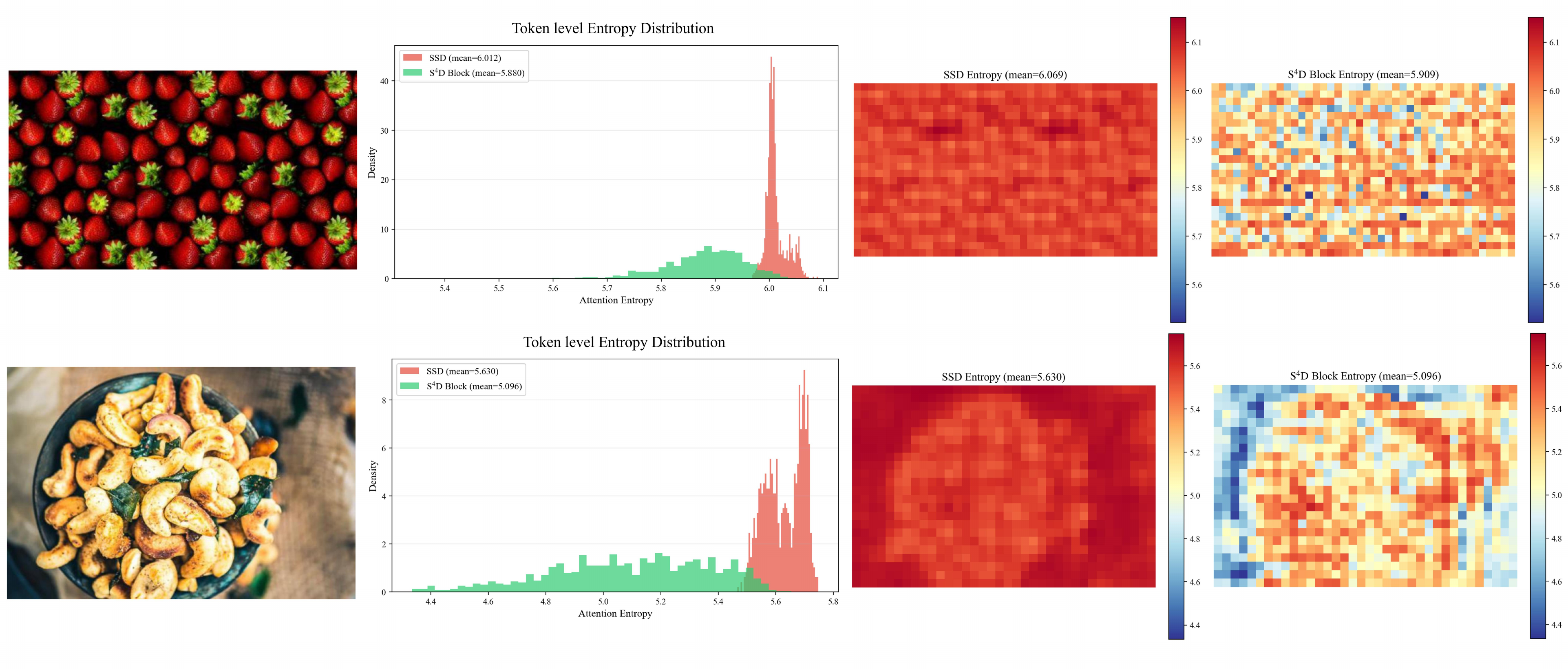}
    \caption{Visualization of token-level entropy distributions and spatial entropy maps. Compared with SSD, the proposed $S^4D$ block reduces the entropy of spatial token responses and produces clearer spatial variation, indicating more discriminative local representations and improved sensitivity to fine-grained image structures.}
    \label{fig:entropy_vis}
\end{figure*}

\begin{table*}[t]
\centering
\caption{Computational efficiency comparison of attention blocks ($d{=}512$, $h{=}8$). Trans.: Transformer Encoder (2 layers); Deform.: Deformable Transformer Decoder (6 layers, 900 queries); SSD Block: ours (2 layers).}
\label{tab:flops_benchmark}
\setlength{\tabcolsep}{4pt}
\begin{tabular}{l|c|ccc|ccc|ccc}
\toprule
\multirow{2}{*}{Resolution} & \multirow{2}{*}{Tokens} 
& \multicolumn{3}{c|}{FLOPs (G)} 
& \multicolumn{3}{c|}{GPU Latency (ms)} 
& \multicolumn{3}{c}{CPU Latency (ms)} \\
& 
& Trans. & Deform. & Ours 
& Trans. & Deform. & Ours 
& Trans. & Deform. & Ours \\
\midrule
$320{\times}320$   & 400  & 2.85  & 24.40 & \textbf{3.06}  & \textbf{0.98}  & 6.73  & 3.26  & \textbf{24.85}  & 262.42 & 38.57  \\
$480{\times}640$   & 1200 & 10.51 & 25.66 & \textbf{9.18}  & 3.68  & 6.74  & \textbf{3.86}  & 109.43 & 265.54 & \textbf{104.53} \\
$640{\times}640$   & 1600 & 15.33 & 26.29 & \textbf{12.25} & 5.97  & 6.87  & \textbf{4.77}  & 175.11 & 269.79 & \textbf{142.14} \\
$800{\times}800$   & 2500 & 28.55 & 27.71 & \textbf{19.14} & 13.15 & 7.07  & \textbf{7.84}  & 368.27 & 279.92 & \textbf{220.43} \\
$1024{\times}1024$ & 4096 & 60.17 & 30.22 & \textbf{31.35} & 31.05 & \textbf{7.31}  & 13.45 & 880.13 & \textbf{318.58} & 423.81 \\
\midrule
Params (M) & — & 6.30 & 23.72 & 7.51 & — & — & — & — & — & — \\
\bottomrule
\end{tabular}
\end{table*}
\newpage
\section{Experiment}
We train MambaCount on the REC-8K training set and evaluate it on the corresponding test split for the REC task.
For the class-agnostic counting task, we train our model using only the training set of FSC-147. The trained model is then evaluated on the FSC-147 test set, which contains object categories different from those in the training set, as well as on the test set of CARPK.

\subsection{Dataset Details}
FSC-147 is a few/zero-shot object counting dataset consisting of 6,135 images covering 147 object categories. Each image is annotated with point-level annotations indicating the locations of object instances. In addition, three exemplar bounding boxes are provided to specify the counting category. 
The dataset contains over 200k object instances with significant variations in scale, density, and appearance, making it highly challenging for generalized counting methods.

CARPK is a widely used dataset for counting cars in parking lot scenes. It contains 1,448 aerial images captured from drone views, with 89,777 annotated cars. Each car is labeled with a bounding box.
The dataset is commonly used to evaluate object counting and detection performance in dense parking environments. The dataset is typically split into 989 training images and 459 testing images.

REC-8K is a referring expression counting dataset designed for language-guided visual counting. It contains 8,011 images with 286,621 point annotations and 7,122 image–referring-expression pairs. Each pair associates an image with a natural language description specifying the target objects to be counted. 
The dataset provides diverse attribute descriptions such as color, location, action, and orientation, enabling evaluation of models on language-driven counting tasks under dense object scenarios.

\begin{table}[t]
\centering
\caption{Ablation study on different prompt configurations and Top-$K$ selection.}
\label{tab:prompt_ablation}
\resizebox{0.9\linewidth}{!}{
\begin{tabular}{l|cc|cc}
\hline
\multirow{2}{*}{Prompt Setting} & \multicolumn{2}{c|}{Val Set} & \multicolumn{2}{c}{Test Set} \\
\cline{2-5}
 & MAE $\downarrow$ & RMSE $\downarrow$ & MAE $\downarrow$ & RMSE $\downarrow$ \\
\hline

Text Only & 13.61 & 56.45 & 13.36 & 107.10 \\

\hline
All-16 & 14.19 & 53.89 & 13.16 & 113.69 \\
All-32 & 13.76 & 54.76 & 13.70 & 113.70 \\

\hline
Top-$K$=4 & \textbf{13.18} & \textbf{53.58} & 12.68 & \textbf{105.05} \\
Top-$K$=8 & 13.60 & 54.32 & 13.71 & 116.29 \\
Top-$K$=16 & 13.22 & 53.62 & \textbf{12.23} & 105.69 \\
Top-$K$=32 & 13.48 & 57.21 & 13.62 & 112.92 \\

\hline
Without Text (Top-$K$=16) & 13.46 & 53.53 & 13.59 & 108.42 \\

\hline
\end{tabular}
}
\end{table}

\subsection{Effect of prompt configuration and Top-$K$ selection}
Table~\ref{tab:prompt_ablation} analyzes the impact of different prompt configurations and Top-$K$ selection strategies. Using only the text prompt provides a reasonable baseline performance (Test MAE 13.36), indicating that textual guidance already offers useful semantic cues for counting. Introducing prototype-based prompts further improves the results, demonstrating that visual prototypes can complement textual semantics.
We further compare different Top-$K$ selection strategies. Using a fixed number of prototypes across all scales (All-16 and All-32) leads to limited improvements, suggesting that uniformly selecting prototypes may introduce redundant or noisy regions. In contrast, selecting Top-$K$ prototypes for each scale yields better performance. We adopt Top-$K$=16 per scale as the default, which achieves the best Test MAE (12.23); other choices ($K$=4/8/32) bring no consistent gain, indicating that selecting an appropriate per-scale $K$ provides a good balance between semantic coverage and noise suppression.
Finally, removing the text prompt (“Without Text”) degrades the performance compared with the full model (13.59 vs. 12.23 MAE), confirming that textual guidance plays an important role in aligning visual features with the target category. Overall, these results demonstrate that combining text prompts with appropriately selected multi-scale prototypes leads to the most effective counting performance.

\begin{table}[t]
\centering
\caption{Comparison of different fusion modules.}
\label{tab:fusion_module}
\resizebox{1\linewidth}{!}{
\begin{tabular}{l|cc|cc}
\hline
\multirow{2}{*}{Fusion Module} & \multicolumn{2}{c|}{Val set} & \multicolumn{2}{c}{Test set} \\
\cline{2-5}
 & MAE $\downarrow$ & RMSE $\downarrow$ & MAE $\downarrow$ & RMSE $\downarrow$ \\
\hline
Transformer \cite{MHSA} & 13.92 & 60.35 & 15.72 & 125.57 \\
Mamba \cite{mamba} & 13.50 & 55.86 & 15.69 & 116.83 \\
MambaOut \cite{mambaout} & 13.85 & 56.91 & 14.56 & 108.94 \\
MLLA \cite{mlla} & 14.12 & 57.26 & 16.87 & 134.68 \\
VSSD \cite{vssd} & 13.91 & 56.08 & 13.32 & 107.08 \\
GLA \cite{GLA} & \textbf{13.03} & \textbf{51.51} & 14.91 & 113.75 \\
\hline
$S^4D$ Block & 13.22 & 53.62 & \textbf{12.23} & \textbf{105.69} \\
\hline
\end{tabular}
}
\end{table}

\subsection{Attention/ Mamba module comparison.}
Table~\ref{tab:fusion_module} compares different fusion mechanisms for vision-language interaction. While Transformer and Mamba-based variants provide moderate improvements, the proposed $S^4D$ block achieves the best performance on the test set (12.23 MAE). This indicates that the structured state-space modeling captures long-range dependencies more effectively than conventional attention-based fusion.

\section{Visualization}
\subsection{Visualization of $S^4D$ Block.}
Fig~\ref{fig:entropy_vis} further analyzes the unconstrained high entropy of spatial token responses within SSD \cite{mamba2}. 
We observe that the original SSD tends to produce excessively high-entropy responses, meaning that attention over spatial tokens is overly diffuse and insufficiently focused on informative regions. 
After introducing our optimization strategy, the excessive entropy in spatial token responses is effectively reduced, leading to more concentrated and semantically meaningful activations. 
The corresponding entropy histograms and spatial maps show that token responses become more structured, with clearer emphasis on object-relevant areas and less interference from irrelevant background regions.
\section{Mamba Details}
State Space Models (SSMs)~\cite{mamba} describe a continuous linear dynamical system that maps an input signal $x(t)$ to an output $y(t)$ through a latent state $h(t)$:

\begin{equation}
h'(t) = \mathbf{A}h(t) + \mathbf{B}x(t), \quad y(t) = \mathbf{C}h(t)
\end{equation}

where $\mathbf{A} \in \mathbb{R}^{N \times N}$ is the state transition matrix, and $\mathbf{B} \in \mathbb{R}^{N \times 1}$ and $\mathbf{C} \in \mathbb{R}^{1 \times N}$ are input and output projection matrices, respectively. After discretization with step size $\Delta$, the system can be written in a recurrent form as:

\begin{equation}
h_t = \bar{\mathbf{A}}h_{t-1} + \bar{\mathbf{B}}x_t, 
\quad 
y_t = \mathbf{C}h_t,
\end{equation}

where $\bar{\mathbf{A}} = \exp(\Delta \mathbf{A})$ and $\bar{\mathbf{B}}$ is the discretized input matrix.

When applied to vision tasks, the input sequence $x_t$ corresponds to visual tokens obtained by flattening spatial feature maps. However, existing SSM-based vision methods typically rely on multi-directional scanning strategies to convert two-dimensional spatial features into one-dimensional sequences. Such scanning operations often disrupt spatial locality and may introduce redundant dependencies, limiting their effectiveness in modeling complex visual structures.

Mamba2 further establishes the State Space Duality (SSD) \cite{mamba2}, which shows that the SSM output can be expressed as a matrix multiplication:

$$Y = \mathbf{M} \cdot X, \quad \mathbf{M} = \mathbf{L} \circ \mathbf{C}\mathbf{B}^T$$

where $\mathbf{L} \in \mathbb{R}^{T \times d}$ is a lower-triangular factor that encodes the causal constraint, $\circ$ is the hadamard product, and $d$ is the state dimension. 
\end{document}